\documentclass[letterpaper]{article} 
\usepackage{aaai24}  
\usepackage{times}  
\usepackage{helvet}  
\usepackage{courier}  
\usepackage[hyphens]{url}  
\usepackage{graphicx} 
\usepackage{natbib}  
\usepackage{caption} 
\usepackage{algorithm}
\usepackage{algorithmic}

\usepackage{newfloat}
\usepackage{listings}
\DeclareCaptionStyle{ruled}{labelfont=normalfont,labelsep=colon,strut=off} 
\floatstyle{ruled}
\newfloat{listing}{tb}{lst}{}
\floatname{listing}{Listing}
\usepackage[utf8]{inputenc} 
\usepackage{booktabs}       
\usepackage{amsfonts}       
\usepackage{nicefrac}       
\usepackage{microtype}      
\usepackage[dvipsnames]{xcolor}
\usepackage{bm}
\usepackage{amsmath}
\usepackage{makecell}
\usepackage{graphicx}
\usepackage{subcaption}
\usepackage{tabularx}
\usepackage{adjustbox}
\usepackage{multirow}

\title{Multi-Architecture Multi-Expert Diffusion Models}

\author {
    Yunsung Lee\textsuperscript{\rm 2}\equalcontrib \quad
    JinYoung Kim\textsuperscript{\rm 3}\equalcontrib \quad
    Hyojun Go\textsuperscript{\rm 3}\equalcontrib \\
    Myeongho Jeong\textsuperscript{\rm 4} \quad
    Shinhyeok Oh\textsuperscript{\rm 1} \quad
    Seungtaek Choi\textsuperscript{\rm 4}\footnote{Corresponding author \quad $^{2, 3, 4}$Work done while at Riiid}
}
\affiliations {
    \textsuperscript{\rm 1}Riiid AI Research
    \textsuperscript{\rm 2}Wrtn Technologies
    \textsuperscript{\rm 3}Twelvelabs
    \textsuperscript{\rm 4}Yanolja
    \\
    sung@wrtn.io, seago0828@gmail.com, gohyojun15@gmail.com \\
    myeongho.jeong@yanolja.com, shinhyeok.oh@riiid.co seungtaek.choi@yanolja.com
}

\begin{document}

\maketitle

\newcommand{\todoc}[2]{{\textcolor{#1}{\textbf{#2}}}}
\newcommand{\todoblue}[1]{\todoc{blue}{\textbf{#1}}}
\newcommand{\todored}[1]{\todoc{red}{#1}}
\newcommand{\todobrown}[1]{\todoc{brown}{#1}}
\definecolor{mhc}{rgb}{0,0.5,0.5}
\newcommand{\todomhc}[1]{\todoc{mhc}{#1}}

\newcommand{\jiny}[1]{\todobrown{\textbf{jinyoung:} #1}}
\newcommand{\hist}[1]{\todored{\textbf{seungtaek:} #1}}
\newcommand{\hyojun}[1]{\todoblue{\textbf{hyojun:} #1}}
\newcommand{\mh}[1]{\todomhc{\textbf{myeongho:} #1}}
\newcommand{\yunsung}[1]{\todoc{purple}{\textbf{yunsung:} #1}}

\definecolor{hehe}{rgb}{0, 0.7, 0}
\newcommand{\shinhyeok}[1]{\todoc{hehe}{\textbf{shinhyeok:} #1}}

\begin{abstract}
In this paper, we address the performance degradation of efficient diffusion models by introducing Multi-architecturE Multi-Expert diffusion models (MEME).
We identify the need for tailored operations at different time-steps in diffusion processes and leverage this insight to create compact yet high-performing models.
MEME assigns distinct architectures to different time-step intervals, balancing convolution and self-attention operations based on observed frequency characteristics.
We also introduce a soft interval assignment strategy for comprehensive training.
Empirically, MEME operates 3.3 times faster than baselines while improving image generation quality (FID scores) by 0.62 (FFHQ) and 0.37 (CelebA).
Though we validate the effectiveness of assigning more optimal architecture per time-step, where efficient models outperform the larger models, we argue that MEME opens a new design choice for diffusion models that can be easily applied in other scenarios, such as large multi-expert models. 
\end{abstract}
\section{Introduction}

\begin{figure*}[!t]
    \centering
    \begin{subfigure}{.70\textwidth}
        \centering
        \includegraphics[width=\linewidth]{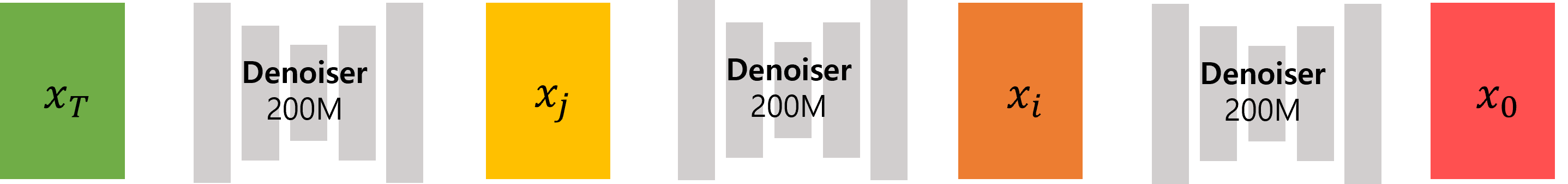}
        \caption{Single large model with an identical architecture}
        \label{subfig:standard_diff_vis}
    \end{subfigure}%

    \begin{subfigure}{.70\textwidth}
        \centering
        \includegraphics[width=\linewidth]{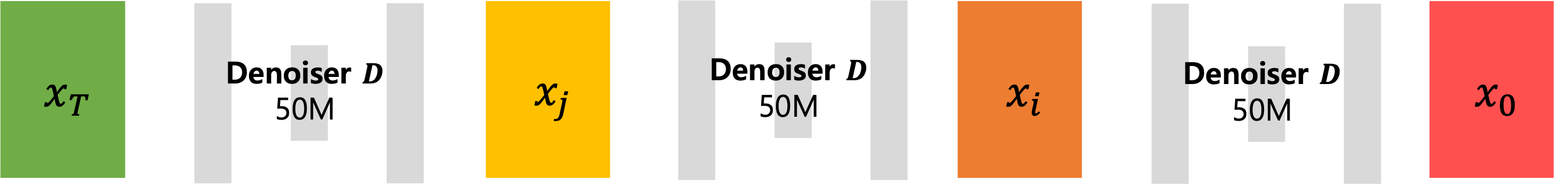}
        \caption{Single small model with an identical architecture}
        \label{subfig:small_diff_vis}
    \end{subfigure}%
    
    
    \begin{subfigure}{.70\textwidth}
        \centering
        \includegraphics[width=\linewidth]{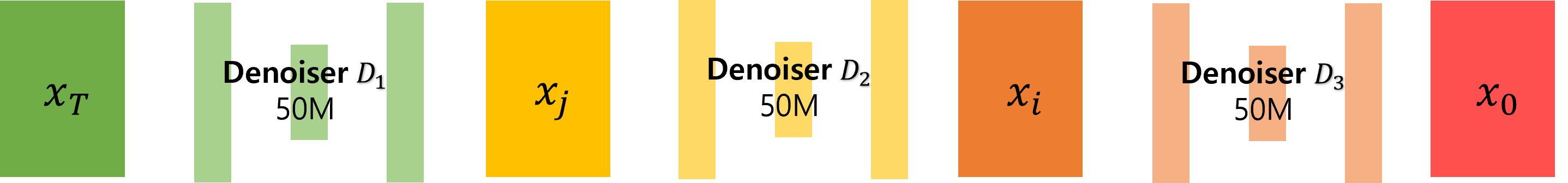}
        \caption{Multiple small expert models with an identical architecture}
        \label{subfig:mult-expert_small_vis}
    \end{subfigure}%
    
    \begin{subfigure}{.70\textwidth}
        \centering
        \includegraphics[width=\linewidth]{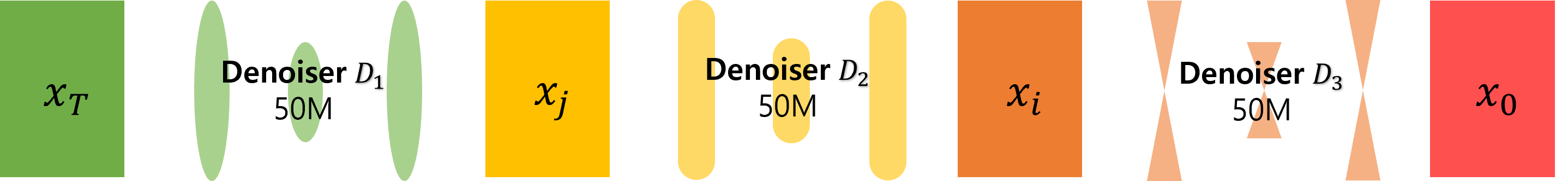}
        \caption{Multiple small expert models with multiple architectures (MEME)}
        \label{subfig:MEME_vis}
    \end{subfigure}
    \caption{\textbf{Comparative illustration of single/multiple-expert models with single/multiple architectures.} Figure~\ref{subfig:standard_diff_vis} depict the standard diffusion models approach, which employs a single large denoiser. 
    To reduce the cost due to the large-scale of the model, a diffusion model with a small denoiser is designed with post-training low-bit or distillation as illustrated in Fig.~\ref{subfig:small_diff_vis}.
    In Fig.~\ref{subfig:mult-expert_small_vis}, to alleviate the performance drop, we consider a configuration with multiple small expert models having identical architectures.
    Finally, our proposed method, the Multi-architecturE Multi-Expert diffusion models (MEME), constructs small expert models with unique optimal architectures for their respective assigned time-step intervals, as visualized in Fig.~\ref{subfig:MEME_vis}. }
    \label{fig:main_schema}
    \vspace{-0.25cm}
\end{figure*}

Diffusion models~\cite{sohl2015deep,song2019generative,ho2020denoising} are a promising approach for generative modeling, and they are likely to play an increasingly important role in diverse domains, including image~\cite{dhariwal2021diffusion,rombach2022high,balaji2022ediffi}, audio~\cite{kong2021diffwave,kim2022guided}, video~\cite{ho2022video,ho2022imagen,zhou2022magicvideo}, and 3D generation~\cite{poole2022dreamfusion,seo2023let}.
However, despite their impressive performance, diffusion models suffer from high computation costs, which stem from the following two orthogonal factors: (i) the lengthy iterative denoising process, and (ii) the cumbersome denoiser networks.
Though there have been several efforts to overcome such limitations~\cite{song2021denoising,bao2022analyticdpm,lu2022dpmsolver,meng2022distillation,song2023consistency}, most of these efforts have focused only on resolving the first factor, such that the cumbersome denoisers still limit their applicability to real-world scenarios.
A few efforts reduce the size of the denoisers based on post-training low-bit quantization~\cite{shang2022post} and distillation~\cite{yang2022diffusion}, as we illustrated in Fig.~\ref{subfig:small_diff_vis}, but they usually achieve such efficiency by compromising on accuracy.

In this paper, we thus aim to build a diffusion model that is compact yet comparable in performance to the large models. For this purpose, we first ask a research question ``\textit{why the traditional diffusion models require such massive parameters?}''. 
\cite{choi2022perception,go2023addressing} suggests that the difficulty of learning diffusion models is that they have to learn all the different tasks at many different time-steps.
One step further, from a frequency perspective, \cite{yang2022diffusion} theoretical explains that diffusion models should learn too many different features in varying time-steps, where diffusion models tend to initially form low-frequency components (e.g., overall image contour) and subsequently fill in high-frequency components (e.g., detailed textures).
However, as they assume the denoiser network to be a linear filter, which is not practical, we aim to investigate empirical evidence to support this claim. 
Specifically, we analyze the per-layer Fourier spectrum for the input $x_t$ at each diffusion time-step $t$, finding that there are significant and consistent variations in the relative log amplitudes of the Fourier-transformed feature maps as $t$ progresses.
This finding indicates that the costly training of large models indeed involves learning to adapt to the different frequency characteristics at each time-step $t$.

One way to leverage this finding is to assign distinct time-step intervals to multiple diffusion models~\cite{go2022towards,balaji2022ediffi}, referred to as the \textit{multi-expert} strategy, in order for models to be specialized in the assigned time-step intervals as shown in Fig.~\ref{subfig:mult-expert_small_vis}. 
However, since \cite{go2022towards} utilized the multi-expert strategy for the conditioned generation with guidance and \cite{balaji2022ediffi} focused on high performance, the architecture efficiency is not considered. 
Most importantly, to the best of our knowledge, previous works have constructed diffusion models with a single architecture, missing the possibility of optimal architectures that better solve the task at each time-step of diffusion.

To this end, we propose to assign different models with \textbf{different architectures} for each different time-step interval, whose base operations vary according to their respective frequency ranges, which we dub \textbf{Multi-architecturE Multi-Expert diffusion models (MEME)} (Fig.~\ref{subfig:MEME_vis}). 
Specifically, we leverage the property that convolutions are advantageous for handling high-frequency components ($t \sim 0$), while multi-head self-attention (MHSA) excels in processing low-frequency components ($t \sim T$)~\cite{d2021convit,dai2021coatnet,park2022how,si2022inception}.
However, a naive hard-shuffling of convolution and MHSA at different time intervals would be suboptimal because the features are inherently a combination of high- and low-frequency components~\cite{d2021convit,si2022inception}.

In order to better adapt to such frequency-specific components, we propose a more flexible denoiser architecture called \textbf{iU-Net}, which incorporates an iFormer~\cite{si2022inception} block that allows for adjusting the channel-wise balance ratio between the convolution operations and MHSA operations. 
We take advantage of the characteristic of diffusion models we discovered that first recover low-frequency components during the denoising process and gradually add high-frequency features.
Consequently, we configure each architecture to have a different proportion of MHSA, effectively tailoring each architecture to suit the distinct requirements at different time-step intervals of the diffusion process.

We further explore methods for effectively assigning focus on specific time-step intervals to our flexible iU-Net. Specifically, we identify a soft interval assignment strategy for the multi-expert models that prefers a soft division over a hard segmentation.
This strategy allows the experts assigned to intervals closer to $T$ to have more chance to be trained with the entire time-step, which prevents excessive exposure to meaningless noises at the time-step $t \sim T$.

Empirically, our MEME diffusion models effectively perform more specialized processing for each time-step interval, resulting in improved performance compared to the baselines.
MEME, with LDM~\cite{rombach2022high} as the baseline, has managed to reduce the computation cost by 3.3 times while training on FFHQ~\cite{karras2019style} and CelebA-HQ~\cite{karras2018progressive} datasets from scratch and has simultaneously improved image generation performance by 0.62 and 0.37 in FID scores, respectively.
By comparing the Fourier-transformed feature maps of MEME and multi-expert with identical architecture, we have confirmed that MEME's multi-architecture approach allows for distinct frequency characteristics suitable for each interval.
Furthermore, MEME not only improves performance when combined with the LDM baseline but also demonstrates successful performance enhancements when integrated with the other diffusion model, DDPM~\cite{ho2020denoising}.

Our main contributions are summarized as follows:
\begin{itemize}
    \item As far as we know, we are the first to identify and address the limitation of diffusion models where vastly different functionalities at each time-step in diffusion processes yield sub-optimal performances.
    \item We propose MEME, a novel diffusion models framework composed of multi-architecture multi-expert denoisers that can balance operations for low- and high-frequency, performing distinct operations for each time-step interval. 
    \item MEME surpasses its large counterparts in terms of generation quality while providing more efficient inference. Trained from scratch on the FFHQ and CelebA datasets, MEME operates \textbf{3.3 times faster} than baselines while \textbf{improving FID} scores by \textbf{0.62} and \textbf{0.37}, respectively. 
\end{itemize}
\section{Related Work}

\subsection{Diffusion Models}

Diffusion models~\cite{sohl2015deep,song2019generative}, a subclass of generative models, generate data through an iterative denoising process. Trained by denoising score-matching objectives, these models demonstrate impressive performance and versatility in various domains, including image~\cite{dhariwal2021diffusion,rombach2022high,balaji2022ediffi}, audio~\cite{kong2021diffwave,popov2021grad,kim2022guided}, video~\cite{ho2022video,zhou2022magicvideo}, and 3D~\cite{poole2022dreamfusion,zeng2022lion,seo2023let} generation. However, diffusion models suffer from significant drawbacks, such as high memory and computation time costs~\cite{kong2021fast,lu2022dpmsolver}. These issues primarily stem from two factors: the lengthy iterative denoising process and the substantial number of parameters in the denoiser model. A majority of studies addressing the computation cost issues of diffusion models have focused on accelerating the iteration process. Among these, \cite{watson2022learning,watson2021learning2,dockhorn2022genie} have employed more efficient differential equation solvers. Other studies have sought to reduce the lengthy iterations by using truncated diffusion~\cite{lyu2022accelerating,zheng2022truncated} or knowledge distillation~\cite{luhman2021knowledge,salimans2022progressive,song2023consistency}. 

In contrast, \cite{shang2022post} and \cite{yang2022diffusion} focus on reducing the size of diffusion models. 
\cite{shang2022post} proposes a post-training low-bit quantization specifically tailored for diffusion models.
\cite{yang2022diffusion} analyzes diffusion models based on frequency, enabling small models to effectively handle high-frequency dynamics by applying wavelet gating and spectrum-aware distillation. 
However, these attempts at lightweight models usually fail to match the performance of large models and rely on resource-intensive training, which assumes the availability of a pretrained diffusion model.
To overcome such dependency, multi-expert strategies have been explored to increase the model capacity~\cite{balaji2022ediffi} while keeping the inference cost at each time step. In this work, our distinction is to enhance the multi-expert strategy, by focusing more on the fact that diffusion models have to learn very different functionality at each time-step, while the existing diffusion models leverages a single operation across the diffusion processes.

\subsection{Combination of Convolutions and Self-attentions}

Since the advent of the Vision Transformer~\cite{dosovitskiy2021an}, there has been active research into why self-attention works effectively in the image domain and how it differs from convolution operations. 
~\cite{park2022how,si2022inception,wang2022antioversmoothing,bai2022improving} explain empirically or theoretically that this is because self-attention operations better capture global features and act as low-pass filters. 
There have been efforts~\cite{dai2021coatnet,d2021convit,park2022how,si2022inception,bu2023towards} aiming to design optimal architectures that better combine the advantages of self-attention and convolution. 
\cite{park2022how,dai2021coatnet} suggest structuring networks with convolution-focused front layers, which are advantageous for high-pass filtering, and self-attention-focused rear layers, which are advantageous for low-pass filtering. 
\cite{d2021convit,si2022inception} propose new blocks that perform operations intermediate between convolution and self-attention. 
Notably, \cite{si2022inception} proposes the iFormer block that allows for adjustable ratios between convolution and self-attention operations. 
While previous efforts have been focused on exploring better operations and architectures for image recognition, we are the first to explore the same questions in diffusion models, revealing the need for diffusion-specific strategies.

\section{Background}

\subsection{Spectrum Evolution over Diffusion Process}
\label{subsec:background_diff}

Diffusion models~\cite{sohl2015deep,song2019generative} work by inverting a stepwise noise process using latent variables. Data points $\mathbf{x}_0$ from the true distribution are perturbed by Gaussian noise with zero mean and $\beta_t$ variance across $T$ steps, eventually reaching Gaussian white noise. As in~\cite{ho2020denoising}, efficiently sampling from the noise-altered distribution $q(\mathbf{x}_t)$ is achieved through a closed-form expression to generate arbitrary time-step $\mathbf{x}_t$:
\begin{align}
& \mathbf{x}_t = \sqrt{\bar{\alpha}_t} \mathbf{x}_0 + \sqrt{1-\bar{\alpha}_t} \bm{\epsilon}, 
\label{eq:diff_x_t}
\end{align}
where $\epsilon \sim \mathcal{N}(0, \mathbf{I})$, $\alpha_{t} = 1 - \beta_{t}$, and $\bar{\alpha}_t=\prod_{s=1}^t \alpha_s$. 

The denoiser, a time-conditioned denoising neural network $\mathbf{s}_{\bm \theta}(\mathbf{x}, t)$ with trainable parameters $\theta$, is trained to reverse the diffusion process by minimizing re-weighted evidence lower bound ~\cite{song2019generative}, as follows: 
\begin{align}
\mathbb{E}_{t,\mathbf{x}_0, \bm\epsilon} \Big[||\nabla{\mathbf{x}_t} \log p(\mathbf{x}_t|\mathbf{x}_0) - \mathbf{s}_{\bm \theta}(\mathbf{x}_t, t) ||_2^2\Big]
\end{align}
In essence, the denoiser learns to recover the gradient that optimizes the data log-likelihood. The previous step data $x_{t-1}$ is generated by inverting the Markov chain:
\begin{align}
\mathbf{x}_{t-1}\leftarrow \frac{1}{\sqrt{1-\beta_t}}(\mathbf{x}_t+\beta_t \mathbf{s}_{\bm \theta}(\mathbf{x}_t, t)) + \sqrt{\beta_t} \bm \epsilon_t
\end{align}
In this reverse process, the insight that diffusion models evolve from rough to detailed was gained through several empirical observations~\cite{ho2020denoising,rombach2022high}.
Beyond them, \cite{yang2022diffusion} provides a numerical explanation of this insight from a frequency perspective by considering the network as a linear filter.
In this case, the optimal filter, known as the Wiener filter~\cite{wiener1949extrapolation}, can be expressed in terms of its spectrum response at every time-step.
Under the widely accepted assumption that the power spectra $\mathbb{E}[|X_0(f)|^2] = A_s(\theta)/f^{\alpha_S(\theta)}$ of natural images $x_0$ follows a power law~\cite{burton1987color,field1987relations,tolhurst1992amplitude} the frequency response of the signal reconstruction filter is determined by the amplitude scaling factor $A_s(\theta)$ and the frequency exponent $\alpha_S(\theta)$.
As the reverse denoising process progresses from $t=T$ to $t=0$, and $\bar{\alpha}$ increases from $0$ to $1$, diffusion models, as analyzed by \cite{yang2022diffusion}, exhibit spectrum-varying behavior over time.
Initially, a narrow-banded filter restores only low-frequency components responsible for rough structures.
As $t$ decreases and $\bar{\alpha}$ increases, more high-frequency components, such as human hair, wrinkles, and pores, are gradually restored in the images.

\subsection{Inception Transformer}

The limitation of transformers in the field of vision is well-known as they tend to capture low-frequency features that convey global information but are less proficient at capturing high-frequency features that correspond to local information~\cite{dosovitskiy2021an,si2022inception}.
To address this shortcoming, \cite{si2022inception} introduced the Inception Transformer, which combines a convolution layer with a transformer, utilizing the Inception module~\cite{szegedy2015going,szegedy2016rethinking,szegedy2017inception}.
To elaborate, the input feature $\mathbf{Z}\in\mathbb{R}^{N\times d}$ is first separated into $\mathbf{Z}_h\in\mathbb{R}^{n\times d_h}$ and $\mathbf{Z}_l\in\mathbb{R}^{n\times d_l}$ along the channel dimension, where $d=d_h+d_l$. 
The iFormer block then applies a high-frequency mixer to $\mathbf{Z}_h$ and a low-frequency mixer to $\mathbf{Z}_l$. Specifically, $\mathbf{Z}_h$ is further split into $\mathbf{Z}_{h_1}$ and $\mathbf{Z}_{h_2}$ along the channel dimension as follows:
\begin{align}
    \mathbf{Y}_{h_1}&=\text{FC}(\text{MP}(\mathbf{Z}_{h_1})), \\
    \mathbf{Y}_{h_2}&=\text{D-Conv}(\text{FC}(\mathbf{Z}_{h_2})),
\end{align}
where $\mathbf{Y}$ denotes the outputs of high-frequency mixer, $\text{FC}$ is fully-connected layer, $\text{MP}$ represents max pooling layer, and $\text{D-Conv}$ is depth-wise convolutional layer.

In the low-frequency mixer, MHSA is utilized to acquire a comprehensive and cohesive representation, as shown in Eq. \ref{eq:low}. This global representation is then combined with the output from the high-frequency mixer as in Eq. \ref{eq:concat}. 
However, due to the potential oversmoothing effect of the upsample operation in Eq. \ref{eq:low}, a fusion module is introduced to counteract this issue and produce the final output, outlined in Eq. \ref{eq:fusion}:
\begin{align}
    \mathbf{Y}_{l}=\text{Up}(\text{MHSA}(\text{AP}(\mathbf{Z}_{h_2}))), \label{eq:low} \\
    \mathbf{Y}_c=\text{Concat}(\mathbf{Y}_{h_1}, \mathbf{Y}_{h_2}, \mathbf{Y}_l), \label{eq:concat} \\
    \mathbf{Y}=\text{FC}(\mathbf{Y}_c+\text{D-Conv}(\mathbf{Y}_c)), \label{eq:fusion}
\end{align}
where $\text{Up}$ denotes upsampling, $\text{AP}$ is average pooling, and $\text{Concat}$ represents concatenation.

\section{Frequency Analysis for Diffusion Models}
\label{sec:freq_analysis}
It is useful to design the architecture with distinct blocks capturing appropriate frequency according to the depth of the block~\cite{d2021convit,touvron2021training}.
In this section, we analyze the frequency-based properties of latents and extracted features according to time-step.

\begin{figure}[!htb]
\centering
\begin{minipage}{\linewidth}
\centering
\begin{tabular}{cccc}
    \begin{subfigure}{0.18\linewidth}
        \includegraphics[width=\linewidth]{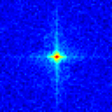}
        \caption{$t=125$}
    \end{subfigure} &
    \begin{subfigure}{0.18\linewidth}
        \includegraphics[width=\linewidth]{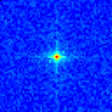}
        \caption{$t=375$}
    \end{subfigure} &
    \begin{subfigure}{0.18\linewidth}
        \includegraphics[width=\linewidth]{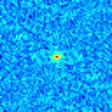}
        \caption{$t=625$}
    \end{subfigure} &
    \begin{subfigure}{0.18\linewidth}
        \includegraphics[width=\linewidth]{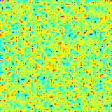}
        \caption{$t=875$}
    \end{subfigure}
\end{tabular}
\vspace{-2mm}
\caption{\textbf{Visualization of the Fourier spectrum of the inputs} used in training diffusion models. As $t$ increases from $0$ to $T$, the high-frequency feature spectrum, initially concentrated towards the center, gradually disappears.}
\label{fig:input_fourier}
\end{minipage}
\end{figure}

\subsection{Frequency Component from Latents.}
From the fact that a Gaussian filter prioritizes the filtering out of high-frequency~\cite{gonzalez2002digital}, it is evident that the training data fed into diffusion models progressively lose their high-frequency spectrum as $t$ increases. 
In Fig. \ref{fig:input_fourier}, we visualize the Fourier spectrum of input latent, output of autoencoder, for training LDM~\cite{rombach2022high} with FFHQ~\cite{karras2019style} dataset.
By illustrating the Fourier coefficients of the periodic function against the corresponding frequency, the training data gradually lose their high-frequency spectrum as $t$ increases. 
It suggests designing a diffusion model that filters different frequency components according to the time-step for dealing with the corresponding features.

\subsection{Frequency Component Focused by Model.}
\begin{figure*}[!t]
    \captionsetup[subfigure]{font=tiny}
    \centering
    \begin{subfigure}{.19\textwidth}
        \centering
        \includegraphics[width=\linewidth]{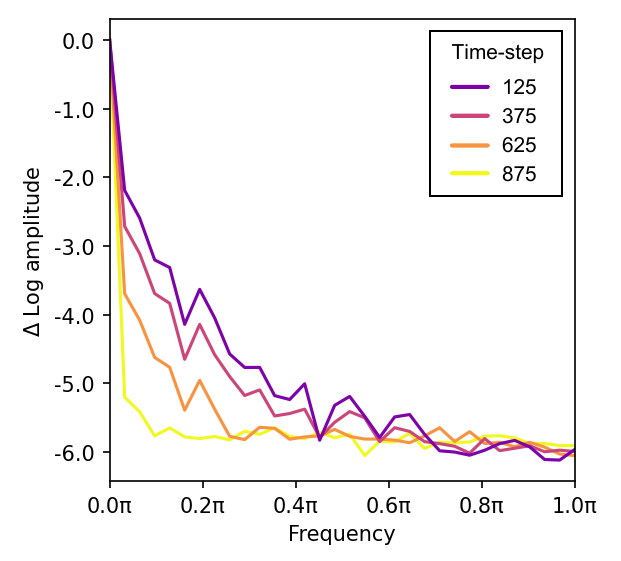}
        \caption{The former layer in the encoder}
    \end{subfigure}%
    \begin{subfigure}{.19\textwidth}
        \centering
        \includegraphics[width=\linewidth]{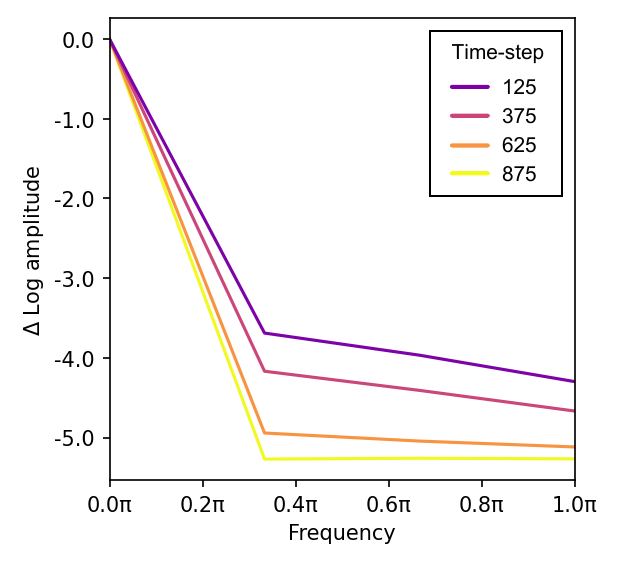}
        \caption{The later layer in the encoder}
    \end{subfigure}%
    \begin{subfigure}{.19\textwidth}
        \centering
        \includegraphics[width=\linewidth]{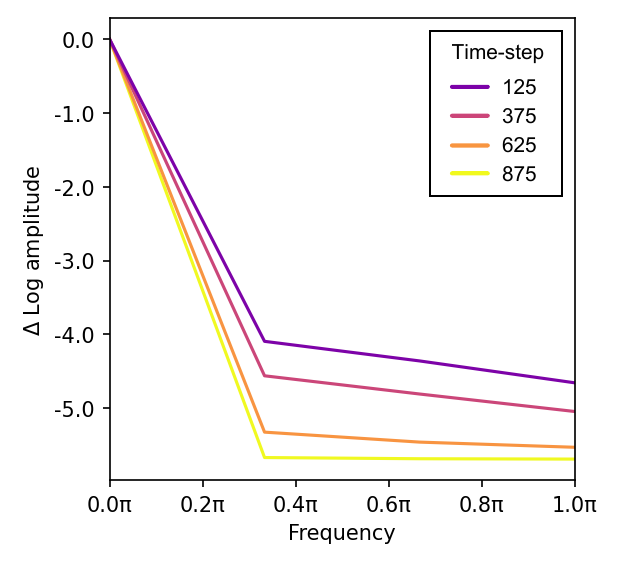}
        \caption{The former layer in the decoder}
    \end{subfigure}%
    \begin{subfigure}{.19\textwidth}
        \centering
        \includegraphics[width=\linewidth]{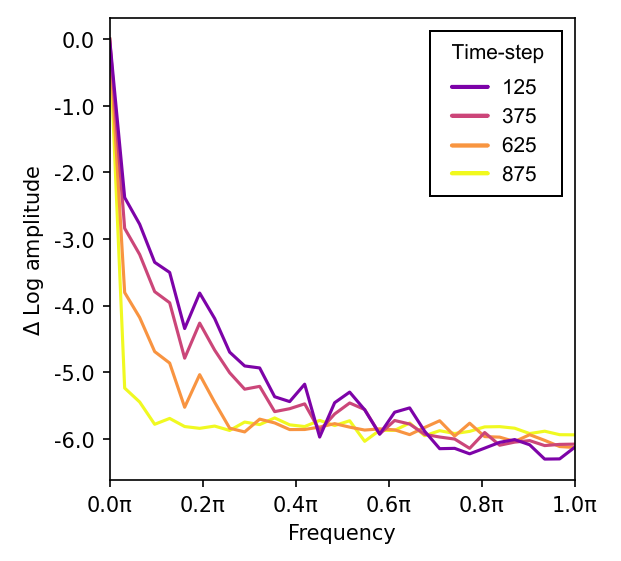}
        \caption{The later layer in the decoder}
    \end{subfigure}
    \caption{\textbf{Visualization of relative log amplitudes} of Fourier transformed feature map obtained from the pre-trained large LDM.
    The $\Delta \text{Log}$ amplitude of high-frequency signals is a difference with log amplitudes at the frequency of $0.0\pi$ and $\pi$.
    We compute it with 10K image samples at each time-step $t\in\{125, 375, 625, 875\}$. 
    It shows the tendency of $\Delta\text{Log}$ amplitude to be interpolated as $t$ is changed.
    In particular, as $t\rightarrow T$, the Fourier transformed features from the model are rapidly changed after $0.0\pi$.}
    \label{fig:ldm_ori_fourier}
    \vspace{-0.3cm}
\end{figure*}

Here, we first introduce the analysis on the frequency for each layer with the distinct depth as~\cite{d2021convit} did. 
We examine the relative log amplitudes of Fourier-transformed feature maps obtained from the pre-trained latent diffusion model (LDM).
\cite{park2022how} reveals that image recognition deep neural networks primarily perform high-pass filtering in earlier layers and low-pass filtering in later layers.
Additionally, in this paper, we further analyze the frequency components focused by the diffusion model with respect to the diffusion time-step.
The captured feature with frequency perspective is illustrated in each subfigure of Fig. \ref{fig:ldm_ori_fourier}, indicating that diffusion models tend to attenuate low-frequency signals more prominently as $t$ increases.
These findings align with the well-established characteristics of a Gaussian filter, known for its tendency to suppress high-frequency components primarily~\cite{gonzalez2002digital}.

\section{Multi-Architecture Multi-Expert}

Based on our frequency analysis for diffusion models, we propose the following significant hypothesis: \textit{By structuring the denoiser model with operations that vary according to each time-step interval, it could potentially enhance the efficiency of the diffusion model's learning process.} 
To validate this hypothesis, two key elements are needed: i) a denoiser architecture with the capacity to adjust the degree of its specialization towards either high or low frequencies, and ii) a strategy for varying the application of this tailored architecture throughout the diffusion process.

\begin{figure}[t]
    \captionsetup[subfigure]{}
    \centering
    \includegraphics[width=0.44\textwidth]{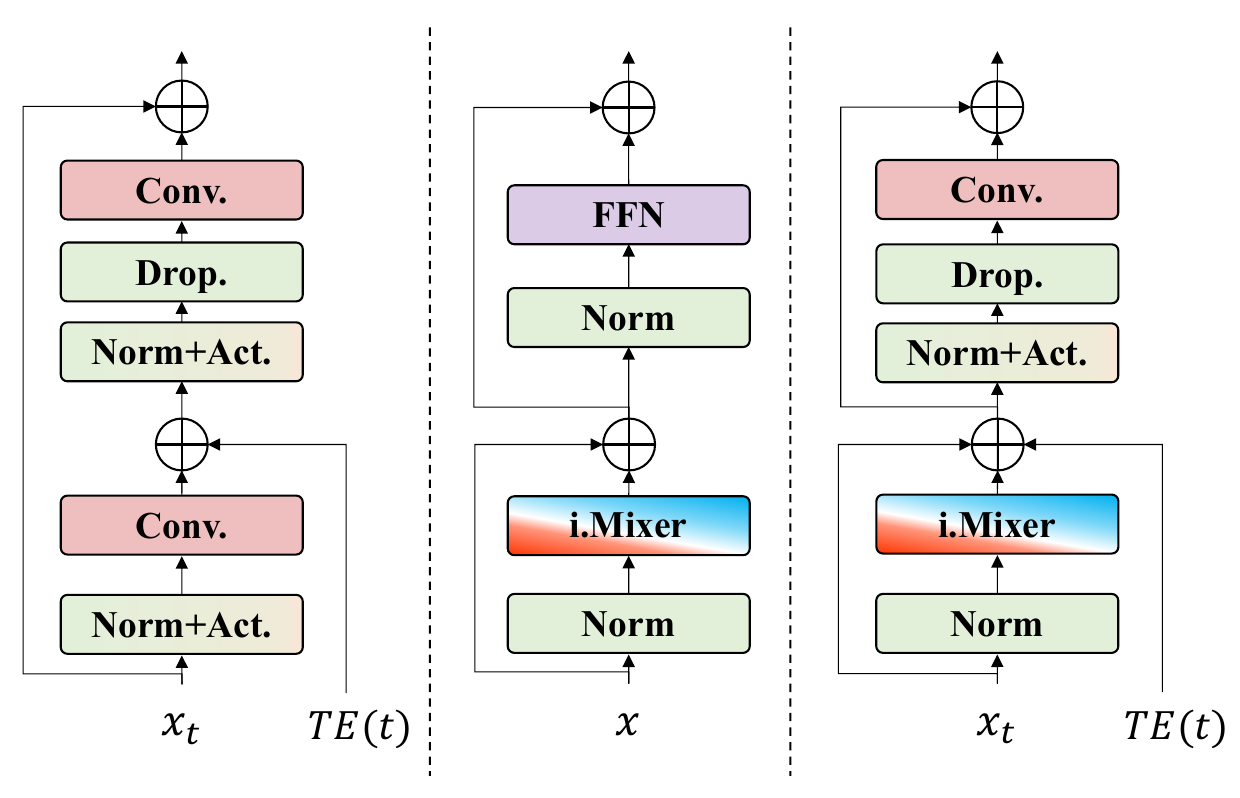}
    \begin{subfigure}{.14\textwidth}
        \centering
        \caption{U-Net}
        \label{subfig:unet}
    \end{subfigure}%
    \begin{subfigure}{.14\textwidth}
        \centering
        \caption{iFormer}
        \label{subfig:iformer}
    \end{subfigure}%
    \begin{subfigure}{.14\textwidth}
        \centering
        \caption{Ours (iU-Net)}
        \label{subfig:iunet}
    \end{subfigure}%
    \caption{\textbf{Comparative illustration of the block in the diffusion models.} The \(\oplus\) denotes element-wise addition and \(TE\) denotes the time-embedding lookup table.} 
    \label{fig:our_block_vis}
    \vspace{-0.2cm}
\end{figure}

\subsection{iU-Net Architecture}

We propose the iU-Net architecture, a variant of U-Net~\cite{ronneberger2015u} that allows for adjusting the ratio of operations favorable to high and low frequencies. 
We utilize a block referred to as the inception transformer (iFormer)~\cite{si2022inception}, which intertwines convolution operations, suitable for high-pass filtering, and Multi-Head Self-Attention (MHSA) operations, suitable for low-pass filtering, with an inception~\cite{szegedy2015going} mixer.
Figure~\ref{fig:our_block_vis} illustrates the manner in which we have adapted the iFormer block to fit the denoiser architecture of diffusion.
This setup allows the iFormer block to regulate the ratio between the convolution-heavy high-frequency mixer and the MHSA-heavy low-frequency mixer in the architecture's composition.
Following \cite{park2022how,dai2021coatnet,wu2021cvt,d2021convit,si2022inception} that tried to combine convolution and MHSA, we set the iU-Net encoder to perform more MHSA operations in the later layers.
We discuss it more technically in a later section.
Furthermore, as in~\cite{cao2022exploring}, rather than completely replacing the block architecture from the U-Net block to the iFormer block, asymmetrically merging the two is effective in constructing an architecture for diffusion model that exploits iFormer.

\begin{table*}[!t]
\renewcommand{\arraystretch}{1.}
\footnotesize
{
    \caption{
    \textbf{Overall Results of Unconditional Generation on FFHQ and CelebA-HQ} We use the Clean-FID implementation to ensure reproducibility. We sample 200 steps using DDIM on the FFHQ, and 50 steps on the CelebA-HQ. Even with $N$ models trained using Multi-Expert and MEME, the total training cost was equivalent to that of a large model. SD is trained through knowledge distillation, which is dependent on having a large pretrained model already, but we can build an efficient model from scratch. The symbols denote $\dagger$: values reported in the original source; $\ddagger$: average value across four architectures; $*$: calculated using checkpoints from our training; $**$: recalculated using pretrained checkpoints from the official repository.
    }
    \vspace{-2mm}
    \centering
    \begin{tabularx}{0.95\textwidth}{X|c|c|c|c|c}
    \hline
    \multicolumn{6}{c}{\textbf{FFHQ $256\times 256$}~~~(DDIM-200)}\\
    \hline
     Model &\#Param$\downarrow$ & MACs$\downarrow$ & FID$\downarrow$ & Prec.$\uparrow$ & Recall $\uparrow$ \\
    \hline
    LDM-L$^{**}$ (635K iter)~\cite{rombach2022high}  & 274.1M & 288.2G & \underline{9.03} & \textbf{0.72} & \underline{0.49} \\
    Lite-LDM$^\dagger$~\cite{yang2022diffusion} & 22.4M & 23.6G & 17.3 & - & - \\
    SD (with Distill.)~\cite{yang2022diffusion} & 21.1M & - & 10.5 & - & - \\
    \hline
    LDM-L$^*$ (540K iter) & 274.1M & 288.2G & 9.14 & \textbf{0.72} & 0.48 \\
    LDM-S$^*$ & 89.5M({\tiny\textcolor{Green}{$ 3.1\times$}}) & 94.2G({\tiny\textcolor{Green}{$3.1\times $}}) & 11.41({\tiny\textcolor{red}{$-2.27$}}) & 0.66({\tiny\textcolor{Red}{$-0.06$}}) & 0.44({\tiny\textcolor{Red}{$-0.04$}}) \\
    iU-LDM-S$^*$ & \textbf{82.6M}({\tiny\textcolor{Green}{$ 3.3\times$}}) & \underline{90.5G}({\tiny\textcolor{Green}{$3.2\times $}}) & 11.64({\tiny\textcolor{red}{$-2.50$}}) & 0.65({\tiny\textcolor{Red}{$-0.07$}}) & 0.45({\tiny\textcolor{Red}{$-0.03$}}) \\
    Multi-Expert$^*$ (w/o Soft) & 89.5M\tiny{$\times 4$}({\tiny\textcolor{Green}{$ 3.1\times$}}) & 94.2G({\tiny\textcolor{Green}{$3.1\times $}}) & 10.42({\tiny\textcolor{red}{$-1.28$}}) & 0.69({\tiny\textcolor{Red}{$-0.03$}}) & 0.46({\tiny\textcolor{Red}{$-0.02$}}) \\
    Multi-Expert$^*$ & 89.5M\tiny{$\times 4$}({\tiny\textcolor{Green}{$ 3.1\times$}}) & 94.2G({\tiny\textcolor{Green}{$3.1\times $}}) & 9.58({\tiny\textcolor{red}{$-0.44$}}) & \underline{0.70}({\tiny\textcolor{Red}{$-0.02$}}) & 0.46({\tiny\textcolor{Red}{$-0.02$}}) \\
    \hline
    MEME$^*$(w/o Soft) & \underline{82.9M}$\ddagger$\tiny{$\times 4$}({\tiny\textcolor{Green}{$ 3.3\times$}}) & \textbf{90.4G$\ddagger$}({\tiny\textcolor{Green}{$3.3\times $}}) & 9.20({\tiny\textcolor{red}{$-0.06$}}) & \underline{0.70}({\tiny\textcolor{Red}{$-0.02$}}) & 0.48({\tiny\textcolor{Green}{$+0.00$}}) \\
    MEME$^*$ & \underline{82.9M}$\ddagger$\tiny{$\times 4$}({\tiny\textcolor{Green}{$ 3.3\times$}}) & \textbf{90.4G$\ddagger$}({\tiny\textcolor{Green}{$3.3\times $}}) & \textbf{8.52}({\tiny\textcolor{Green}{$+0.62$}}) & \textbf{0.72}({\tiny\textcolor{Green}{$+0.00$}}) & \textbf{0.50}({\tiny\textcolor{Green}{$+0.02$}}) \\
    \hline
    \end{tabularx}
    \centering
    \begin{tabularx}{0.95\textwidth}{X|c|c|c|c|c}
    \hline
    \multicolumn{6}{c}{\textbf{CelebA-HQ $256\times 256$}~~~(DDIM-50)}\\
    \hline
    Model &\#Param$\downarrow$ & MACs$\downarrow$ & FID$\downarrow$ & Prec.$\uparrow$ & Recall $\uparrow$ \\
    \hline
    LDM-L$^{**}$ (410K iter)~\cite{rombach2022high} & 274.1M & 288.2G & \underline{5.92} & \underline{0.71} & \textbf{0.49} \\
    Lite-LDM$^\dagger$~\cite{yang2022diffusion} & 22.4M & 23.6G & 14.3 & - & - \\
    SD$^\dagger$ (with Distill.)~\cite{yang2022diffusion} & 21.1M & - & 9.3 & - & - \\
    \hline
    LDM-S$^*$ & 89.5M({\tiny\textcolor{Green}{$ 3.1\times$}}) & 94.2G({\tiny\textcolor{Green}{$3.1\times $}}) & 9.11({\tiny\textcolor{red}{$-3.19$}}) & 0.61({\tiny\textcolor{Red}{$-0.10$}}) & 0.45({\tiny\textcolor{Red}{$-0.04$}}) \\
    iU-LDM-S$^*$ & \textbf{82.6M}({\tiny\textcolor{Green}{$ 3.3\times$}}) & \underline{90.5G}({\tiny\textcolor{Green}{$3.2\times $}}) & 9.06({\tiny\textcolor{red}{$-3.14$}}) & 0.60({\tiny\textcolor{Red}{$-0.11$}}) & 0.47({\tiny\textcolor{Red}{$-0.02$}})\\
    Multi-Expert$^*$ & 89.5M\tiny{$\times 4$}({\tiny\textcolor{Green}{$ 3.1\times$}}) & 94.2G({\tiny\textcolor{Green}{$3.1\times $}}) & 7.00({\tiny\textcolor{red}{$-1.08$}}) & 0.67({\tiny\textcolor{Red}{$-0.04$}}) & \underline{0.48}({\tiny\textcolor{Red}{$-0.01$}})\\
    \hline
    MEME$^*$ & \underline{82.9M}$\ddagger$\tiny{$\times 4$}({\tiny\textcolor{Green}{$ 3.3\times$}}) & \textbf{90.4G}$\ddagger$({\tiny\textcolor{Green}{$3.2\times $}}) & \textbf{5.55}({\tiny\textcolor{Green}{$+0.37$}}) & \textbf{0.73}({\tiny\textcolor{Green}{$+0.02$}}) & \textbf{0.49}({\tiny\textcolor{Green}{$+0.00$}}) \\
    \hline
    \end{tabularx}
    \label{tab:unconditional}
    \vspace{-0.3cm}
}
\end{table*}

\subsection{Multi-Architecture Multi-Expert Strategy}

\label{subsec:meme_strategy}

\textbf{Architecture Design for Experts} 
To facilitate the construction of structures capable of accommodating diverse architectures, we employ a multi-expert strategy~\cite{go2022towards,balaji2022ediffi}, but assign different architectures to each expert according to the frequency component.
In each architecture, the ratio of dimension sizes for high and low channels is defined by two factors: layer depth and diffusion time-step. 
The former is well-known to enable the frequency dynamic feature extraction by focusing on lower frequency as a deeper layer~\cite{park2022how,dai2021coatnet}.
For more technical derivation, let $d^k$ be the channel size in the $k$-th layer, $d_h^k$ be the dimension size for the high mixer, and $d_l^k$ for the low mixer, satisfying $d^k=d_h^k+d_l^k$.  
Based on the analysis in Fig. \ref{fig:ldm_ori_fourier}, the ratio in each iFormer block is defined for dealing with appropriate frequency components according to the depth; $d_h^k/d_l^k$ decreases as a deeper block.
On the other hand, the latter (diffusion time-step) can be associated with the frequency components based on the observation we found; as time-step $t$ increases, the lower frequency components are focused. 
Therefore, we configure the iU-Net architecture such that the ratio of $d_h^k/d_l^k$ decreases faster for the denoiser taking charge of the expert on the larger $t$.

\textbf{Soft Expert Strategy.} As suggested in~\cite{go2022towards}, one of $N$ experts $\Theta_n$ is trained on the uniform and disjoint interval $\mathbb{I}_n=\left\{ t \bigg| t \in \left(\frac{(n-1)}{N}T, \frac{n}{N}T\right] \right\}$ for $n=1,...,N$.
However, for the large $n$, expert $\Theta_n$ takes as noised input images by near Gaussian noise $\epsilon_n\sim\mathcal{N}(\sqrt{\bar{\alpha}_n}x_0,(1-\bar{\alpha}_n)\mathbf{I})$, which makes it challenging for meaningful learning to take place with $\Theta_n$.
To address this, we propose a \textit{soft expert strategy}, where each $\Theta_n$ learns on the interval $\mathbb{I}_n$ with a probability of $p_n$ denoted as the expertization probability\footnote{Note that When $p_n=1$ regardless of $n$, it can be denoted as \textit{hard expert strategy}~\cite{go2022towards}.}.
Otherwise, it learns on the entire interval $\bigcup_{n=1}^N\mathbb{I}_n$ with the remaining probability of $(1-p_n)$.

Since it is evident that $\Theta_n$ for large $n$ takes more noised images, larger $p_n$ as $n\rightarrow N$ is a more flexible strategy for training multi-expert, yielding $p_1\ge\dots\ge p_N$.

\section{Experiments}

In this section, we demonstrate the capability of MEME to enhance the efficiency of diffusion models.
Firstly, we showcase how our model can achieve superior performance over the baseline models, despite being executed with less computation.
Secondly, we verify if our MEME model, as hypothesized, indeed incorporates appropriate Fourier features for each time-step interval input.

We evaluated the unconditional generation of models on two datasets, FFHQ~\cite{karras2019style} and CelebA-HQ~\cite{karras2018progressive}.
We construct models based on the LDM framework.
All pre-trained auto-encoders for LDM were obtained from the official repository\footnote{\url{https://github.com/CompVis/latent-diffusion}}.

MEME employs a multi-expert structure composed of multiple small models, each of which has its channel dimension reduced from 224 to 128.
The use of these smaller models is denoted by appending an `S' to the model name, such as in LDM-S and iU-LDM-S. We set the number of experts $N$ to 4 for all multi-expert settings, including MEME.

All experiments were conducted on a single NVIDIA A100 GPU.
We primarily utilize the AdamW optimizer~\cite{loshchilov2017decoupled}.
The base learning rate is set according to the oigianl LDM~\cite{rombach2022high}.
Notably, our smaller models employ a setting that doubles the batch size, which is not feasible with the original LDM on a single A100.
Correspondingly, the base learning rate for our smaller models is also doubled compared to the standard settings.

We assess the quality of our generated models using the FID score~\cite{heusel2017gans}.
As the FID score can be challenging to replicate due to the settings of the reference set, we calculate it using the publicly available Clean-FID~\cite{parmar2022aliased} implementation\footnote{\url{https://github.com/GaParmar/clean-fid}}.
Particularly for the FFHQ dataset, the availability of a fixed reference set allows for a fair comparison of generation quality across all evaluated generative models on Clean-FID.
To verify the efficiency of our trained model, we compare its model size and computational cost using the number of parameters and Multiply-Add cumulation (MACs)\footnote{\url{https://github.com/sovrasov/flops-counter.pytorch}} as metrics. We provide detailed configurations and hyperparameters regarding the models in the Appendix.

\label{subsec:uncond_results}

\begin{figure*}[t]
    \centering
    \begin{tabular}{@{}c@{\hspace{0.5mm}}c@{\hspace{0.5mm}}c@{\hspace{0.5mm}}c@{\hspace{0.5mm}}c@{\hspace{0.5mm}}c}
        
        \raisebox{0.05\height}{\rotatebox{90}{Baseline \tiny{(LDM)}}} 
        \adjincludegraphics[clip,width=0.115\textwidth,trim={0 0 0 0}]{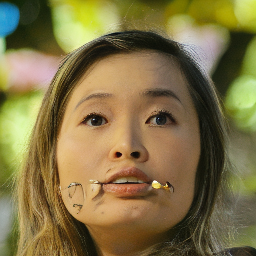} &
        \adjincludegraphics[clip,width=0.115\textwidth,trim={0 0 0 0}]{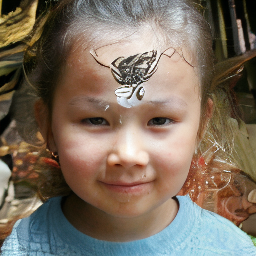} &
        \adjincludegraphics[clip,width=0.115\textwidth,trim={0 0 0 0}]{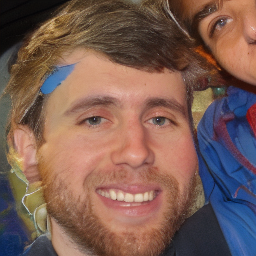} &
        \adjincludegraphics[clip,width=0.115\textwidth,trim={0 0 0 0}]{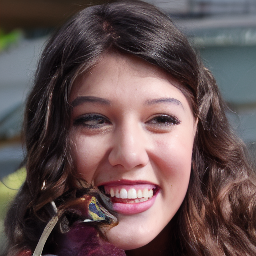} &
        \adjincludegraphics[clip,width=0.115\textwidth,trim={0 0 0 0}]{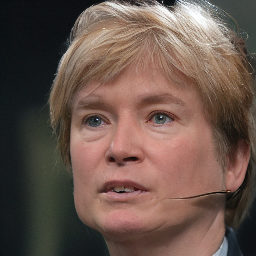} &
        \adjincludegraphics[clip,width=0.115\textwidth,trim={0 0 0 0}]{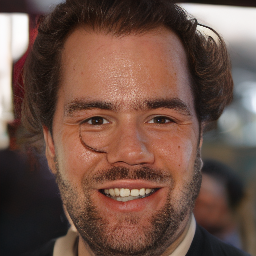} \\
        
        \raisebox{0.0\height}{\rotatebox{90}{MEME \tiny{($3.3\times$ fast)}}} 
        \adjincludegraphics[clip,width=0.115\textwidth,trim={0 0 0 0}]{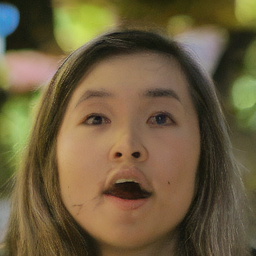} &
        \adjincludegraphics[clip,width=0.115\textwidth,trim={0 0 0 0}]{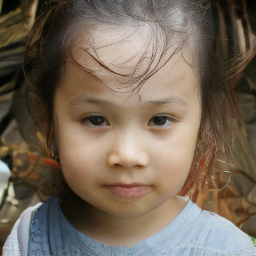} &
        \adjincludegraphics[clip,width=0.115\textwidth,trim={0 0 0 0}]{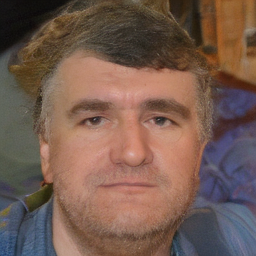} &
        \adjincludegraphics[clip,width=0.115\textwidth,trim={0 0 0 0}]{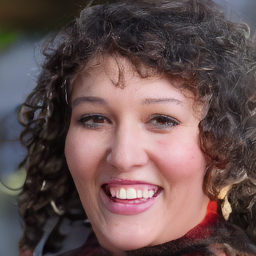} &
        \adjincludegraphics[clip,width=0.115\textwidth,trim={0 0 0 0}]{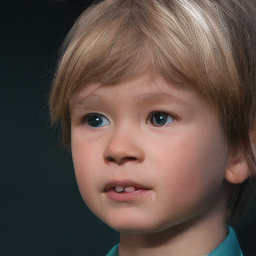} &
        \adjincludegraphics[clip,width=0.115\textwidth,trim={0 0 0 0}]{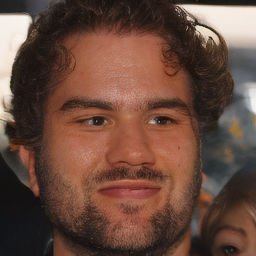} \\
    \end{tabular}
    \vspace{-2mm}
    \caption{\textbf{Samples from baseline LDM-L and MEME trained on FFHQ.} The baseline often generates unnatural aspects in images, whereas our approach MEME shows fewer such cases. }
    \label{fig:qualitative_analysis}
    \vspace{-0.2cm}
\end{figure*}

\noindent\textbf{Image Generation Results}~
The results of our model trained on FFHQ~\cite{karras2019style} and CelebA-HQ~\cite{karras2018progressive} datasets are shown in Table~\ref{tab:unconditional}. 
Despite requiring only 3.3 times less computation cost (MACs), our model demonstrates an improvement in performance (FID). Specifically, we observe a gain of 0.62 in FID for FFHQ and 0.4 in FID for CelebA.
In the case of MEME and Multi-Expert, they require $N$ models to be loaded into system memory for inference. 
However, in large-scale sample inference scenarios, it is possible to load only one expert into system memory while storing intermediate outputs on the disk, yielding less cost to the inferring process. 
Our approach allows for an improvement of 3.3 times in memory cost, which is equivalent to that of a single denoiser. In our experimental setting with $N = 4$, even if all experts are loaded into system memory for inference, it only requires an additional 20.9\% of memory compared to the single large model. 
Further details regarding these two inference scenarios can be found in the Appendix. 
It is also worth noting that in our experiments, the four experts of Multi-Expert and MEME incurred less than 30\% of the computation time compared to LDM-L based on A100 GPU. 
Therefore, the overall training cost requires less than an additional 20\% of resources. 
The qualitative results illustrated in Fig. \ref{fig:qualitative_analysis} show that the generated images by our methods are superior to those by baseline.
Further experiments on other domains (i.e., ImageNet) are in Appendix.

\begin{figure*}[!t]
    \captionsetup[subfigure]{font=tiny}
    \centering
    \begin{subfigure}{.19\textwidth}
        \centering
        \includegraphics[width=\linewidth]{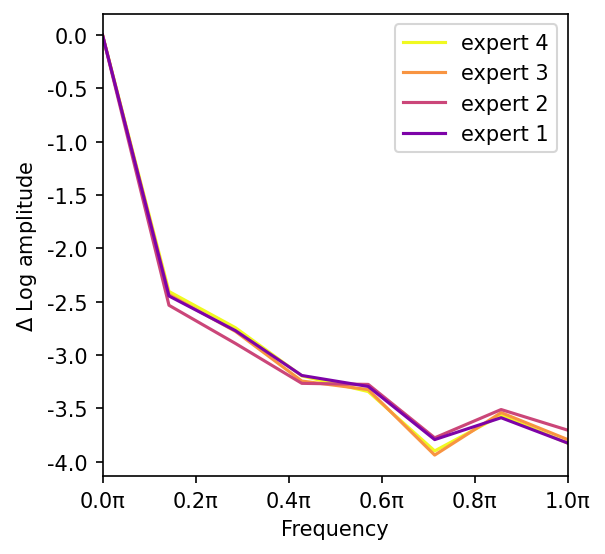}
        \caption{Multi-Expert, $t=125$}
    \end{subfigure}%
    \begin{subfigure}{.19\textwidth}
        \centering
        \includegraphics[width=\linewidth]{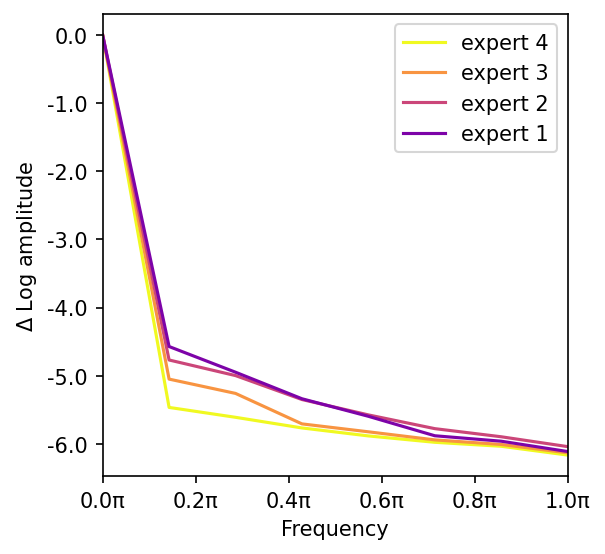}
        \caption{MEME, $t=125$}
    \end{subfigure}%
    \begin{subfigure}{.19\textwidth}
        \centering
        \includegraphics[width=\linewidth]{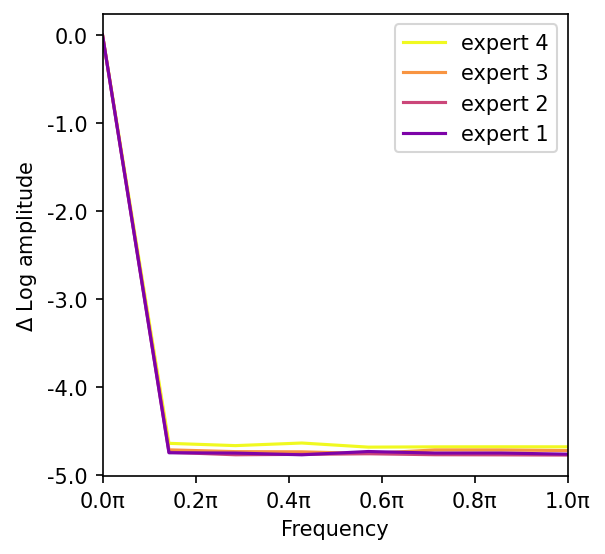}
        \caption{Multi-Expert, $t=875$}
    \end{subfigure}%
    \begin{subfigure}{.19\textwidth}
        \centering
        \includegraphics[width=\linewidth]{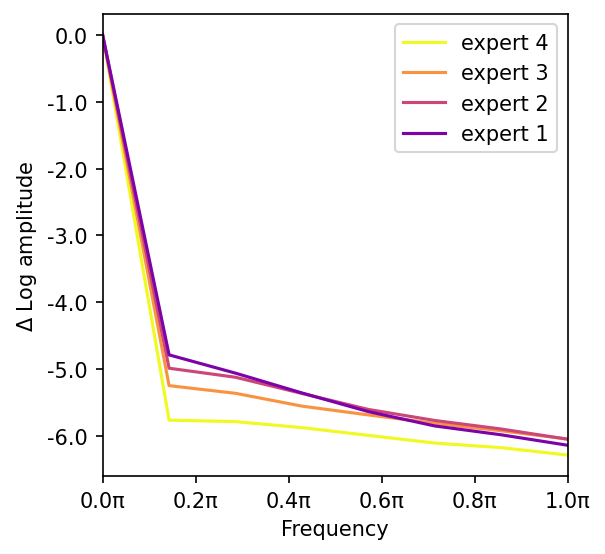}
        \caption{MEME, $t=875$}
    \end{subfigure}%
    \vspace{-2mm}
    \caption{
    \textbf{Fourier Analysis Comparison between Multi-Expert and MEME} 
    Even with the same input $t$, we can confirm that MEME exhibits different characteristics for each expert. 
    MEME demonstrates a similar trend as the pre-trained large model shown in Fig. \ref{fig:ldm_ori_fourier}, where experts responsible for intervals close to $t=T$ rapidly reduce high frequencies. 
    In contrast, Multi-Expert composed of the same architecture shows that the frequency characteristics of features for the same time-step input are not significantly distinguished from each other.}
    \label{fig:meme_fourier}
    \vspace{-0.4cm}

\end{figure*}

\noindent\textbf{Module Ablation}~
Table \ref{tab:unconditional} provides ablation study for various methods on FFHQ dataset. 
Firstly, when training with the baseline LDM-S, which involves standard diffusion training, performance drop (FID -2.27, -2.50 for LDM-S, iU-LDM-S, respectively) occurs. 
Although the incorporation of Multi-Expert mitigates the performance drop to some extent, there is still a degradation (FID -1.28) compared to the baseline LDM-L. 
In contrast, MEME not only reduces computational cost through the use of smaller-sized denoisers but also achieves performance improvement (FID +0.62).

Additionally, we found that the soft-expert strategy is more efficient than the hard-expert strategy, where each expert focuses on its designated region. 
We empirically discovered that a strategy for training the multi-expert with the expertization probability denoted as $p_n$ is beneficial. 
We configured the probabilities: $p_1 = 0.8$, $p_2 = 0.4$, $p_3 = 0.2$, and $p_4 = 0.1$. 
Different configurations for $p_n$ are provided in the Appendix.

\noindent\textbf{Fourier Analysis of MEME}~
\label{subsec:meme_fourier}
In this section, unlike the analysis shown in Fig.~\ref{fig:ldm_ori_fourier}, we investigate whether the experts in MEME possess the ability to capture the corresponding frequency characteristics that are advantageous for their respective intervals as illustrated in Fig.~\ref{fig:meme_fourier}.
MEME, composed of various architectures, exhibits different characteristics for each expert; experts responsible for intervals closer to $t=T$ quickly reduce high frequencies. 
In contrast, the Multi-Expert, composed of the same architecture, fails to significantly differentiate the frequency characteristics of features when the same time-step input is provided. 
Particularly for $t=875$, which requires the ability to capture low-frequency components, it is difficult to distinguish the features of all experts.

\noindent\textbf{MEME on Top of the Other Diffusion Baseline}~
\label{subsec:meme_on_ddpm}
\begin{table}[!t]
\renewcommand{\arraystretch}{1.}
\footnotesize
{
    \centering
    \vspace{-0.25cm}
    \begin{tabular}{c|c|c}
    \hline
    \multicolumn{3}{c}{\textbf{CelebA $64\times 64$}}\\
    \hline
    Model & \#Param$\downarrow$ & FID$\downarrow$ \\
    \hline
    ADM-S & 90M & 49.56 \\
    iU-ADM-S & \textbf{82M}({\tiny\textcolor{Green}{$ 1.1\times$}}) & 50.08({\tiny\textcolor{red}{$-0.52$}}) \\
    Multi-Expert & 90M \tiny{$\times 4$} & 47.29({\tiny\textcolor{Green}{$+2.27$}}) \\
    MEME & \textbf{82M} \tiny{$\times 4$}({\tiny\textcolor{Green}{$ 1.1\times$}}) & \textbf{43.09}({\tiny\textcolor{Green}{$+6.47$}})\\
    \hline
    \end{tabular}
    \caption{\textbf{Results when applied to ADM baseline} MEME outperforms ADM-S baseline~\cite{choi2022perception}.}
    \label{tab:model_ddpm}
    \vspace{-4mm}
}
\end{table}

In order to explore the generalizability of MEME, we adopted the experimental setup used for architecture validation in \cite{choi2022perception}. We trained a lightweight version of ADM~\cite{dhariwal2021diffusion} (referred to as ADM-S) on the CelebA-64 dataset with batch size 8 and 200K iterations. The FID measurement was conducted from 10K samples from DDIM~\cite{song2021denoising} with 50 steps. The results demonstrate that our MEME exhibits effective performance (FID +6.47) even in the context of ADM. Furthermore, the consistent trend is in line with the results observed in the LDM experiments.

\section{Conclusion}
In this paper, we studied the problem of improving efficient diffusion models, with the distinction of adopting multiple architectures to suit the specific frequency requirements at different time-step intervals. 
By incorporating the iU-Net architecture, we provide a more flexible and efficient solution for handling the complex distribution of frequency-specific components across time-steps. 
Our experiments validated that the proposed method, named \textbf{MEME}, outperforms existing baselines in terms of both generation performance and computational efficiency, making it a more practical solution for real-world applications.
While we confirm that assigning optimal architectures per time-step results in efficient models outperforming larger ones, we believe MEME offers a new design choice for diffusion models.

\bibliography{aaai24}

\clearpage

\appendix
\newpage

\section*{\Large Appendix: Multi-Architecture Multi-Expert Diffusion Models}


\section{Experimental Details}

In this section, we provide the details of experiments in Section 6. All experiments are conducted with a single A100 GPU.

\subsection{Experimental Details for LDM Baseline}
\label{subsec:ldm_hparams}

Commonly used hyperparameters of the models involved in image generation experiments on FFHQ~\cite{karras2019style} and CelebA~\cite{karras2018progressive} datasets are presented in Table.~\ref{tab:supp_hyperparams}. 
Additionally, the hyperparameters specific to our expert models within the Multi-architecturE Multi-Expert (MEME) framework, employing the iU-Net, are outlined in Table.~\ref{tab:supp_meme_hparams}. 
Our implementation for the models in experiments is based on the official Latent Diffusion Models (LDM; \citealt{rombach2022high}) repository\footnote{\url{https://github.com/CompVis/latent-diffusion}}.

In the FFHQ experiments, all models generate 50K samples for evaluation via the Denoising Diffusion Implicit Models (DDIM; \citealt{song2021denoising}) with a 200-step sampling process. The Fréchet Inception Distance (FID; \citealt{heusel2017gans}) is computed utilizing the Clean-FID~\cite{parmar2022aliased} official code\footnote{\url{https://github.com/GaParmar/clean-fid}}, with the entire 70K FFHQ dataset serving as the reference image set. The constancy of this reference set bolsters the reproducibility of the Clean-FID computations. The LDM-L model, trained for 635K iterations, is incorporated with pretrained weights obtained from the LDM~\cite{rombach2022high} official repository. In contrast, another model is independently trained for 540K iterations to provide a comparable measure against Multi-Expert and MEME, specifically in terms of GPU memory and time costs. For both Multi-Expert and MEME, the batch size is set twice that of the large model, equating the GPU memory costs when using a single A100 GPU. This configuration leads to an approximate usage of 50GB VRAM in the given system. The process of sequentially training four small experts, each for 135k iterations on a single A100 GPU, exhibits similar time costs to training a large model for 520K iterations. 

For experiments involving the CelebA-HQ~\cite{karras2018progressive} dataset, all models are subject to a DDIM 50-step sampling process to generate 50K samples. To align experimental settings with the FFHQ dataset, the Clean-FID score is computed with the entire 30K CelebA-HQ dataset employed as the reference image set.

\begin{table*}[thbp]

\caption{Hyperparameters for the LDMs producing the numbers shown in Table.~1. All models are trained on a single NVIDIA A100. Further details for iU-LDM-S and MEME architectures are shown in \ref{tab:supp_meme_hparams}}
\begin{center}
\begin{adjustbox}{max width=.8\textwidth}
\begin{tabular}{lcccc}
\toprule
& \multicolumn{2}{c}{Large} & \multicolumn{2}{c}{Small} \\
\cmidrule(lr){2-3} \cmidrule(lr){4-5}
& CelebA-HQ $256 \times 256$ & FFHQ $256 \times 256$ & CelebA-HQ $256 \times 256$ & FFHQ $256 \times 256$  \\
\midrule
$f$ & 4 & 4 & 4 & 4 \\
$z$-shape & $64 \times 64 \times 3$ & $64 \times 64 \times 3$ &  $64 \times 64 \times 3$ & $64 \times 64 \times 3$\\
$\vert \mathcal{Z} \vert$ & 8192 & 8192 & 8192 & 8192  \\
Diffusion steps &1000 & 1000 &1000 & 1000 \\
Noise Schedule & linear & linear & linear & linear \\
$N_{\text{params}}$ & 274M & 274M & 89.5M & 89.5M \\
Channels & 224 & 224 & 128 & 128 \\
Depth & 2 & 2 & 2 & 2 \\
Channel Multiplier & 1,2,3,4 & 1,2,3,4 & 1,2,3,4 & 1,2,3,4 \\
Attention resolutions & 32, 16, 8 & 32, 16, 8 & 32, 16, 8 & 32, 16, 8 \\
Head Channels & 32 & 32 & 32 & 32 \\
Batch Size & 48 & 42 & 96 & 84 \\
Iterations$^*$ & 410k & 520k, 635k & 85k & 135k \\
Learning Rate& $\text{8.4e-5}$ & $\text{9.6e-5}$ & $\text{1.68e-4}$ & $\text{1.92e-4}$\\
\bottomrule
\end{tabular}

\end{adjustbox}
\end{center}
\label{tab:supp_hyperparams}
\end{table*}

\begin{table*}[!t]
    \begin{center}
    \caption{Configurations of the proposed expert models with iU-Net. $\dagger$ denotes the architecture used for iU-LDM-S}
    \label{tab:supp_meme_hparams}

    \resizebox{1.0\linewidth}{!}{
        \begin{tabular}{c|c|c|c|c|c}
        \toprule 
        Stage & Layer & expert$^\dagger$ $\Theta_1$ & expert $\Theta_2$ & expert $\Theta_3$ & expert $\Theta_4$ \\
        \midrule
        \multirow{4}{*}{enc\#1} & \makecell[c]{iFormer \\ Block} & \makecell[c]{$\begin{bmatrix} 3 \times 3, \mathrm{stride}~1, 128 \\ \begin{Bmatrix} d_h/d = 3/4 \\ d_l/d = 1/4 \\ \mathrm{pool~stride}~2 \end{Bmatrix} \end{bmatrix} \times 2$} & \makecell[c]{$\begin{bmatrix} 3 \times 3, \mathrm{stride}~1, 128 \\ \begin{Bmatrix} d_h/d = 3/4 \\ d_l/d = 1/4 \\ \mathrm{pool~stride}~2 \end{Bmatrix} \end{bmatrix} \times 2$}& \makecell[c]{$\begin{bmatrix} 3 \times 3, \mathrm{stride}~1, 128 \\ \begin{Bmatrix} d_h/d = 3/4 \\ d_l/d = 1/4 \\ \mathrm{pool~stride}~2 \end{Bmatrix} \end{bmatrix} \times 2$}& \makecell[c]{$\begin{bmatrix} 3 \times 3, \mathrm{stride}~1, 128\\ \begin{Bmatrix} d_h/d = 3/4 \\ d_l/d = 1/4 \\ \mathrm{pool~stride}~2 \end{Bmatrix} \end{bmatrix} \times 2$}\\
        \cmidrule{2-6}
        ~ & \makecell[c]{Res \\ Block} & $ 3 \times 3, \mathrm{stride}~1, 128 $ & $ 3 \times 3, \mathrm{stride}~1, 128 $ & $ 3 \times 3, \mathrm{stride}~1, 128 $  & $ 3 \times 3, \mathrm{stride}~1, 128 $\\
        \midrule
        \multirow{4}{*}{enc\#2} & \makecell[c]{iFormer \\ Block} & \makecell[c]{$\begin{bmatrix} 3 \times 3, \mathrm{stride}~1, 256 \\ \begin{Bmatrix} d_h/d = 5/8 \\ d_l/d = 3/8 \\ \mathrm{pool~stride}~2 \end{Bmatrix} \end{bmatrix} \times 2$} & \makecell[c]{$\begin{bmatrix} 3 \times 3, \mathrm{stride}~1, 256 \\ \begin{Bmatrix} d_h/d = 1/2 \\ d_l/d = 1/2 \\ \mathrm{pool~stride}~2 \end{Bmatrix} \end{bmatrix} \times 2$}& \makecell[c]{$\begin{bmatrix} 3 \times 3, \mathrm{stride}~1, 256 \\ \begin{Bmatrix} d_h/d = 3/8 \\ d_l/d = 5/8 \\ \mathrm{pool~stride}~2 \end{Bmatrix} \end{bmatrix} \times 2$}& \makecell[c]{$\begin{bmatrix} 3 \times 3, \mathrm{stride}~1, 256\\ \begin{Bmatrix} d_h/d = 1/4 \\ d_l/d = 3/4 \\ \mathrm{pool~stride}~2 \end{Bmatrix} \end{bmatrix} \times 2$}\\
        \cmidrule{2-6}
        ~ & \makecell[c]{Res \\ Block} & $ 3 \times 3, \mathrm{stride}~1, 256 $ & $ 3 \times 3, \mathrm{stride}~1, 256 $ & $ 3 \times 3, \mathrm{stride}~1, 256 $  & $ 3 \times 3, \mathrm{stride}~1, 256 $\\
        \midrule
        \multirow{4}{*}{enc\#3} & \makecell[c]{iFormer \\ Block} & \makecell[c]{$\begin{bmatrix} 3 \times 3, \mathrm{stride}~1, 384 \\ \begin{Bmatrix} d_h/d = 1/2 \\ d_l/d = 1/2 \\ \mathrm{pool~stride}~2 \end{Bmatrix} \end{bmatrix} \times 2$} & \makecell[c]{$\begin{bmatrix} 3 \times 3, \mathrm{stride}~1, 384 \\ \begin{Bmatrix} d_h/d = 3/8 \\ d_l/d = 5/8 \\ \mathrm{pool~stride}~2 \end{Bmatrix} \end{bmatrix} \times 2$}& \makecell[c]{$\begin{bmatrix} 3 \times 3, \mathrm{stride}~1, 384 \\ \begin{Bmatrix} d_h/d = 1/4 \\ d_l/d = 3/4 \\ \mathrm{pool~stride}~2 \end{Bmatrix} \end{bmatrix} \times 2$}& \makecell[c]{$\begin{bmatrix} 3 \times 3, \mathrm{stride}~1, 384\\ \begin{Bmatrix} d_h/d = 1/8 \\ d_l/d = 7/8 \\ \mathrm{pool~stride}~2 \end{Bmatrix} \end{bmatrix} \times 2$}\\
        \cmidrule{2-6}
        ~ & \makecell[c]{Res \\ Block} & $ 3 \times 3, \mathrm{stride}~1, 384 $ & $ 3 \times 3, \mathrm{stride}~1, 384 $ & $ 3 \times 3, \mathrm{stride}~1, 384 $  & $ 3 \times 3, \mathrm{stride}~1, 384 $\\
        \midrule
        \multirow{1}{*}{enc\#4} & \makecell[c]{iFormer \\ Block} & \makecell[c]{$\begin{bmatrix} 3 \times 3, \mathrm{stride}~1, 512 \\ \begin{Bmatrix} d_h/d = 1/4 \\ d_l/d = 3/4 \\ \mathrm{pool~stride}~2 \end{Bmatrix} \end{bmatrix} \times 2$} & \makecell[c]{$\begin{bmatrix} 3 \times 3, \mathrm{stride}~1, 512 \\ \begin{Bmatrix} d_h/d = 1/8 \\ d_l/d = 7/8 \\ \mathrm{pool~stride}~2 \end{Bmatrix} \end{bmatrix} \times 2$}& \makecell[c]{$\begin{bmatrix} 3 \times 3, \mathrm{stride}~1, 512 \\ \begin{Bmatrix} d_h/d = 1/16 \\ d_l/d = 15/16 \\ \mathrm{pool~stride}~2 \end{Bmatrix} \end{bmatrix} \times 2$}& \makecell[c]{$\begin{bmatrix} 3 \times 3, \mathrm{stride}~1, 512\\ \begin{Bmatrix} d_h/d = 1/16 \\ d_l/d = 15/16 \\ \mathrm{pool~stride}~2 \end{Bmatrix} \end{bmatrix} \times 2$}\\
        \midrule
        \multicolumn{2}{c|}{\#Param. (M) }  & 82.56 & 82.85 & 83.11 & 83.32 \\
        \bottomrule
        \end{tabular}
    }
    \end{center}
\end{table*}

\subsection{Experimental Details for ADM-S Baseline}

In an effort to investigate the potential for employing MEME within the scope of pixel-level diffusion models other than LDM, we incorporate the specific experimental configurations previously utilized by~\cite{choi2022perception}. These configurations were originally devised for the purpose of validating p2-weighting~\cite{choi2022perception} model architecture. In this particular experimental context, we have implemented the small ADM model~\cite{dhariwal2021diffusion}, termed ADM-S, equipped with a total of 90 million parameters. Fundamentally, our experimental model implementation is based on the official repository\footnote{\url{https://github.com/jychoi118/P2-weighting}} of~\cite{choi2022perception}. The hyperparameters pertaining to this experiment can be found in Table.~\ref{table:supp_adm_hparams}.

\begin{table*}[!t]
\centering
\caption{Hyperparameters for the ADM-S~\cite{dhariwal2021diffusion} producing the numbers shown in Table.~2. All models are trained on a single NVIDIA A100. $\ddagger$ denotes: average value across four architectures.}
\begin{tabular}{lcccc}
\hline
               & ADM-S                   & iU-ADM-S                & Multi-Expert            & MEME              \\ \hline
$T$            & 1000                    & 1000                    & 1000                    & 1000                    \\
$\beta_t$      & linear                  & linear                  & linear                  & linear                  \\
Model Size     & 90                      & 82                      & 90 $\times 4$           & 82$^\ddagger$ $\times 4$                     \\
Channels       & 128                     & 128                     & 128                     & 128                     \\
Blocks         & 1                       & 1                       & 2                       & 2                       \\
Self-attn      & bottle                  & bottle                  & bottle                  & bottle              \\
Heads Channels & 64                      & 32                      & 64                      & 64                      \\
BigGAN Block   & yes                     & yes                     & yes                     & yes                     \\
Dropout        & 0.1                     & 0.1                     & 0.1                     & 0.1                     \\
Learning Rate  & $\text{2e}^{\text{-5}}$ & $\text{2e}^{\text{-5}}$ & $\text{2e}^{\text{-5}}$ & $\text{2e}^{\text{-5}}$ \\
Images (M)     & 1.6                     & 1.6                     & 1.6                     & 1.6                     \\ 
\hline
\end{tabular}
\label{table:supp_adm_hparams}
\end{table*}

\subsection{Practical Benefits and Limitations}
\label{subsec:practical_beneift}

The foremost benefit derived from the multi-expert strategy is the considerable reduction in computational time costs. This mirrors the empirical observations made by~\cite{balaji2022ediffi}, who found that within a practical setting, the total inference speed of the model does not vary with the number of experts, $N$. The inference speed stays constant with increasing $N$, as it's defined by the average model size of the experts.

However, a potential limitation of this approach lies in the total memory requirements. If all the expert models are loaded into the GPU memory at once, the memory requirements will increase proportionally with the number of experts, $N$. Despite this, the multi-expert strategy for diffusion models provides a memory-efficient alternative that can reduce the memory requirements at the expense of model loading time. Specifically, one can first load only the single expert model responsible for inference at an early time-step into the GPU memory. Then, after storing the intermediate outputs in disk memory, the current expert model on GPU memory is replaced with the next expert model to resume the diffusion process. It would drastically offer memory savings while providing still reasonable inference times than that of large models.


\section{The Effects of Expertization Probability in Soft-Expert}

\begin{table}[!t]
\renewcommand{\arraystretch}{1.}
\footnotesize
{
    \centering
    \begin{tabular}{c|c|c}
    \hline
    \multicolumn{3}{c}{\textbf{FFHQ $256\times 256$}}\\
    \hline
    Expert Strategy & Multi-Expert & MEME \\
    \hline
    {[1.0, 1.0, 1.0, 1.0]} & 10.42 & 9.20 \\
    {[0.6, 0.6, 0.6, 0.6]} & 10.13 & 9.08 \\
    {[0.8, 0.4, 0.2, 0.1]} & \textbf{9.58} & \textbf{8.52} \\
    \hline
    \end{tabular}
    \caption{FID values on the FFHQ dataset depending on how the expertization probability is assigned. The `Expert Strategy' column represents [$p_1, p_2, p_3, p_4$]. In this context, [1.0, 1.0, 1.0, 1.0] denotes the hard-expert~\cite{go2022towards}, [0.6, 0.6, 0.6, 0.6] denotes the soft-expert with a constant expertization probability, and [0.8, 0.4, 0.2, 0.1] denotes the soft-expert with decreasing expertization probabilities.}
    \label{tab:supp_soft_ablation}
}
\end{table}

In establishing the soft-expert strategy, we hypothesize that experts dealing with more highly diffused inputs suffer from learning meaningful semantics. Therefore, we posit that setting the expertization probabilities, $p_n$, to decrease as $n$ increases ($p_1\ge\dots\ge p_N$) would be more effective than maintaining them all constant. We conducted an experimental comparison to test this hypothesis, contrasting hard-expert~\cite{go2022towards}, soft-expert with constant $p_n$, and soft-expert with decreasing $p_n$.

The results of this comparative study are presented in Table.~\ref{tab:supp_soft_ablation}. The elements in the Expert Strategy column represent [$p_1, p_2, p_3, p_4$]. Thus, [1.0, 1.0, 1.0, 1.0] denotes the hard-expert~\cite{go2022towards}, [0.6, 0.6, 0.6, 0.6] denotes the soft-expert with a constant expertization probability, and [0.8, 0.4, 0.2, 0.1] denotes the soft-expert with decreasing expertization probability. The training and evaluation process details followed those outlined in the manuscript.

\section{The Effects of The Number of Experts}

\begin{table}[!htb]
    \centering
    \renewcommand{\arraystretch}{1.}
    \footnotesize
    \begin{tabular}{c|c|c}
        \hline
        \multicolumn{3}{c}{\textbf{FFHQ $256\times 256$}}\\
        \hline
        & \#Expert $N$ & FID \\
        \hline
        \multirow{3}{*}{LDM-S Multi-Expert}&$N=2$ & 10.97 \\
        &$N=4$ & 10.42 \\
        &$N=6$ & 10.28  \\
        \hline
    \end{tabular}
    \caption{Changes in FID according to the number of experts $N$ in LDM-S Multi-Expert training. All results were trained under the hard-expert expertization probability setting.}
    \label{tab:supp_n_ablation}
\end{table}


As noted in the manuscript, the total inference time in the multi-expert strategy does not increase as the number of experts $N$ increases. 
However, the cost of GPU memory may increase proportionally with $N$. To understand the performance difference based on the number of experts, we varied the number of experts in a multi-expert setting with hard expertization probability and measured the FID~\cite{heusel2017gans} value.

In Table.~\ref{tab:supp_n_ablation}, we show the FID values obtained from training the multi-expert model with different values of $N$: 2, 4, and 6. The performance improvement from $N=2$ to $N=4$ is substantial, but there is not a large increase from $N=4$ to $N=6$. Hence, we have set $N=4$ as our default value. The details of the training and evaluation process followed those outlined in the manuscript.

\section{ImageNet Experiments}

\begin{table*}[htbp]
\centering
\begin{tabular}{l|c|c|c|c|c|c|c}
\hline
 & Inception Score $\uparrow$ & FID $\downarrow$ & sFID $\downarrow$ & Precision $\uparrow$ & Recall $\uparrow$ & \#Param $\downarrow$ & MACs $\downarrow$ \\
\hline
Baseline (LDM-L) & 108.93 & \textbf{13.17} & 19.23 & 0.76 & 0.61 & 395M & 411G \\
\hline
\textbf{MEME} & \textbf{109.95} & 13.19 & \textbf{19.18} & \textbf{0.76} & \textbf{0.61} & \textbf{103M} $\times$ 4 & \textbf{114G} \\
\hline
\end{tabular}
\caption{Comparison of MEME with the LDM-L model trained on Imagenet for 400K iterations}
\label{tab:supp_imagenet}
\end{table*}

For validation on large-scale datasets beyond the face image domain, we conducted an experiment on the ImageNet dataset. Table \ref{tab:supp_imagenet} shows a comparison of MEME with the LDM-L model trained on Imagenet for 400K iterations. The baseline LDM model follows the original configuration of \cite{rombach2022high}. We performed the evaluation on 10K samples. MEME performs comparatively or surpasses the baseline LDM model but with a much more efficient number of parameters and MACs.

\section{Experimental Details for Fourier Analysis}

Following \cite{park2022how}, we analyze the feature maps in the Fourier space to confirm what frequency the pretrained large diffusion models focus on at each time-step (as shown in Fig. 3), or to verify that our proposed MEME learns the characteristics of the responsible intervals more effectively than a single architecture multi-expert (as shown in Fig. 6). 

We perform a Fourier transformation on the feature maps, converting them into a two-dimensional frequency domain. These converted feature maps are then represented in a normalized frequency domain, where the highest frequency components correspond to $f = {-\pi, +\pi}$, while the lowest frequency components coincide with $f = 0$. To enhance the clarity of our visualizations, we focus on presenting only the half-diagonal components. The features required for conducting Fourier analysis are calculated based on the average of features derived from 10,000 randomly sampled input data from the FFHQ~\cite{karras2019style} dataset.

The approach of \cite{park2022how} involves visualizing the $\Delta$ log amplitude for all layers within a single model. However, in our analysis, we incorporate multi-expert diffusion models which add two additional dimensions: the time-step and expert models. This necessitates a different visualization approach, where instead of plotting values corresponding to multiple layers within one model, we chart values per time-step (Fig. 3), or alternatively, per expert model (Fig. 6). This enables a more nuanced understanding of how each time-step is handled across the denoising process, or how each expert model performs.

\section{Societal Impacts}

Generative models, including diffusion models, have the potential to significantly impact society, particularly in the context of DeepFake applications~\cite{franks2018sex} and biased data~\cite{jain2022imperfect,torralba2011unbiased}. 
One of the key concerns lies in the amplification of misinformation and the erosion of trust in visual media. 
Furthermore, if generative models are trained on biased data or intentionally manipulated content, they can inadvertently perpetuate and exacerbate social biases~\cite{jain2022imperfect}, leading to the dissemination of misleading information and the manipulation of public perception.

\section{Limitations}

Our research highlights the significance of customizing architectural designs to align with the specific timestep of diffusion models. 
In order to achieve this, our primary focus lies in tuning the operations within each layer through the modulation of the mixing ratio between convolution and self-attention.
However, there are two limitations that can be addressed in future work.

Firstly, our research recognizes the yet unexplored territory of determining the optimal mixing ratio between convolution and self-attention. Although we demonstrate that increased convolution leads to enhanced performance in latent spaces with lower noise, the precise optimization of the mixing ratio remains a task yet to be accomplished, similar to the advancements achieved through neural architecture search~\cite{zoph2017neural}.
To address this, introducing a neural architecture scheme that adapts to varying timesteps can hold significant potential for advancing diffusion models.

Secondly, our research does not delve into exploring other architectural design factors, such as pooling techniques.
Future work can focus on investigating the impact of different pooling techniques on the performance of diffusion models. 
Additionally, exploring the combination of convolution, self-attention, and other architectural elements, such as residual connections~\cite{he2016deep} or skip connections~\cite{ronneberger2015u}, could provide further insights into optimizing the overall architecture for diffusion models.

\section{Qualitative Resutls}

\begin{figure*}[htbp]
\centering
	\setlength{\tabcolsep}{0pt}
	\begin{tabular}{c}
	\toprule
	Random samples on the CelebA-HQ dataset \\
	\midrule
	\includegraphics[width=0.95\textwidth]{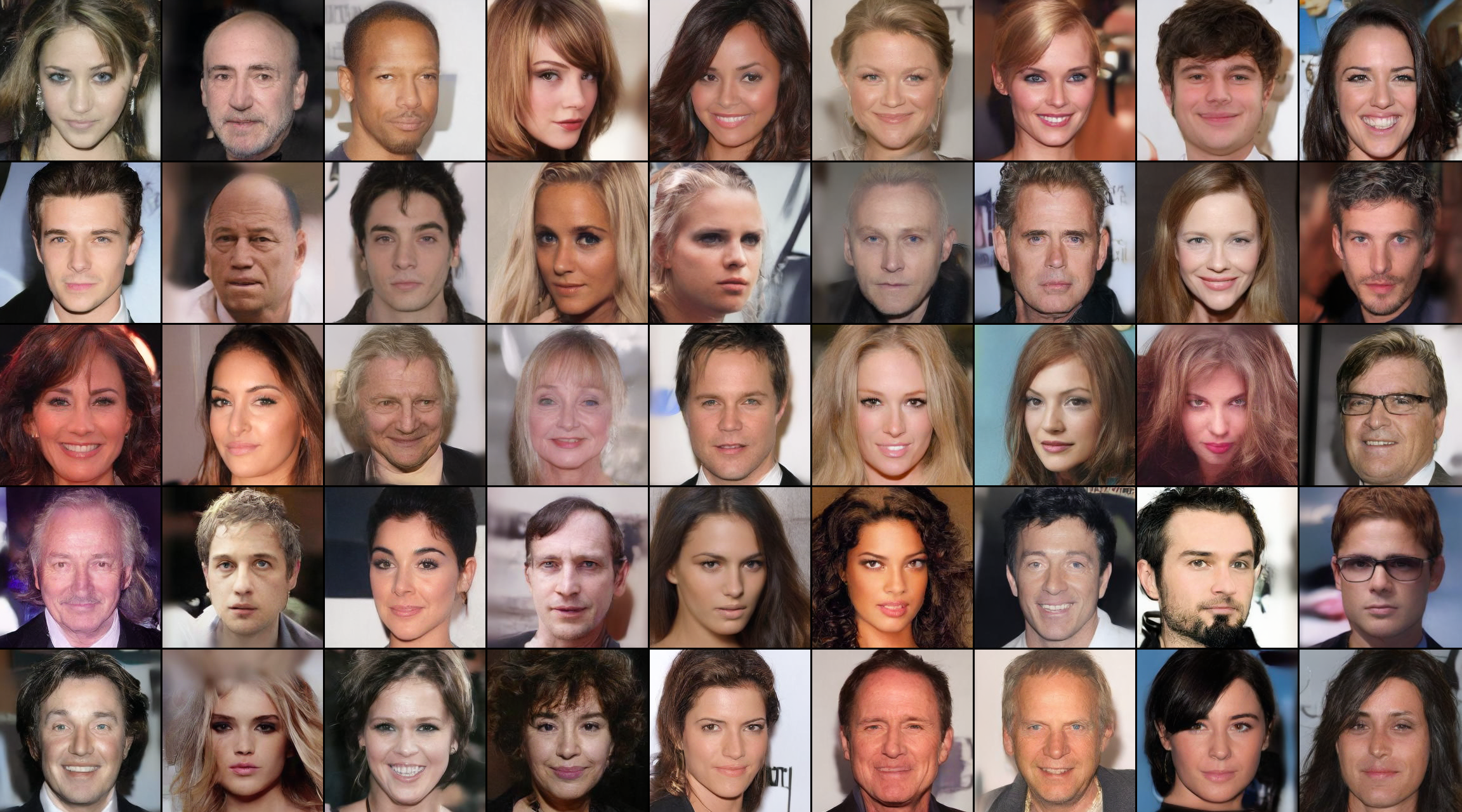}\\
	\bottomrule
	\end{tabular}
 
\caption{Random samples of our MEME on the CelebA-HQ dataset. Sampled with 50 DDIM steps and $\eta=0$.}
\label{fig:supp_qualitative_celeba}
\end{figure*}

\begin{figure*}[htbp]
\centering
	\setlength{\tabcolsep}{0pt}
	\begin{tabular}{c}
	\toprule
	Random samples on the FFHQ dataset \\
	\midrule
	\includegraphics[width=0.95\textwidth]{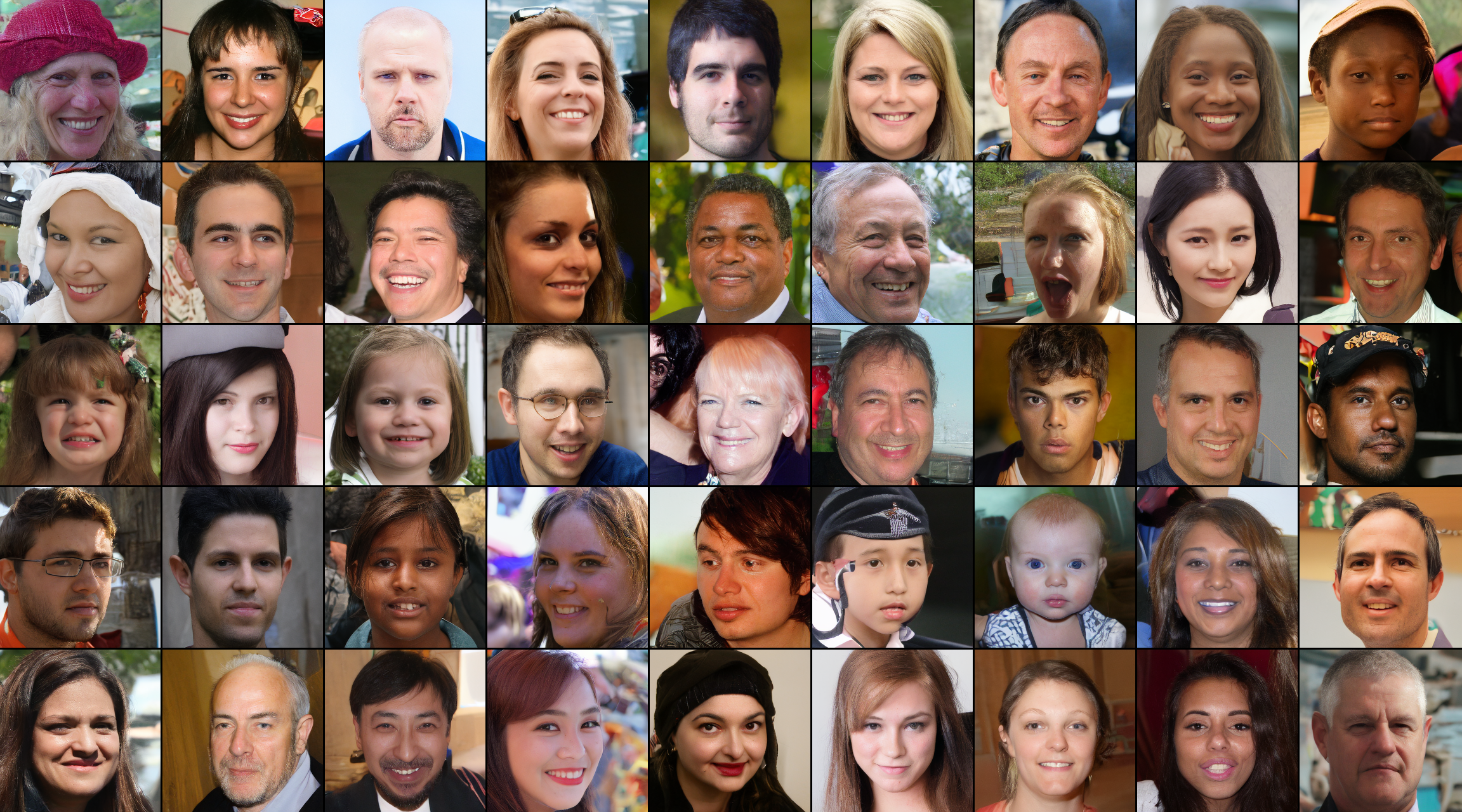}\\
	\bottomrule
	\end{tabular}

\caption{Random samples of MEME on the FFHQ dataset. Sampled with 200 DDIM steps and $\eta=1$.}
\label{fig:supp_qualitative_ffhq}
\end{figure*}

We provide additional qualitative results for our MEME models for the CelebA-HQ~\cite{karras2018progressive}, and FFHQ datasets (Fig.~\ref{fig:supp_qualitative_celeba} - \ref{fig:supp_qualitative_ffhq}).

\end{document}